\RequirePackage{fix-cm}
\documentclass{svjour3}                     
\smartqed  
\usepackage{graphicx}
\usepackage{caption}
\usepackage{subcaption}
\usepackage{array}
\usepackage{amsmath}
\usepackage{amssymb}
\usepackage{amsfonts}
\usepackage{bm}
\usepackage{xcolor}
\usepackage{multirow,array}
\usepackage{comment}
\usepackage{url} 
\usepackage{tikz}
\usepackage{pgfplots}
\usepackage[normalem]{ulem}
\usetikzlibrary{matrix,positioning}
\usepackage{makecell}
\usepackage[flushleft]{threeparttable}
\usepackage{multicol}

\def\makeheadbox{{%
\hbox to0pt{\vbox{\baselineskip=10dd\hrule\hbox
to\hsize{\vrule\kern3pt\vbox{\kern3pt
\hbox{This is a pre-peer-review version of this article.}
\kern3pt}\hfil\kern3pt\vrule}\hrule}%
\hss}}}

\DeclareSymbolFont{bbold}{U}{bbold}{m}{n}
\DeclareSymbolFontAlphabet{\mathbbold}{bbold}

\definecolor{darkyellow}{RGB}{64,64,0}

\definecolor{luisas_color}{HTML}{5c4ae4}
\definecolor{ann_color}{HTML}{42e6f5}

\definecolor{fredrik_color}{HTML}{234567}

\journalname{Autonomous Agents and Multi-Agent Systems}
\begin{document}

\title{Scalar reward is not enough: A response to Silver, Singh, Precup and Sutton (2021)}

\titlerunning{Scalar reward is not enough} 

\author{Peter Vamplew \and Benjamin J. Smith \and Johan K\"allstr\"om \and Gabriel Ramos \and Roxana R\u{a}dulescu \and Diederik M.\ Roijers \and Conor F. Hayes \and  Fredrik Heintz \and Patrick Mannion \and Pieter J.K. Libin \and Richard Dazeley \and Cameron Foale}

\authorrunning{Vamplew, Smith et al.} 
\institute{
            Peter Vamplew, Cameron Foale \at
              Federation University Australia,
              Ballarat, Australia\\
              \email{p.vamplew@federation.edu.au}
              \email{c.foale@federation.edu.au}
            \and
            Benjamin J. Smith \at
             Center for Translational Neuroscience,
              University of Oregon \\
              Eugene, OR,
              United States of America\\
              \email{benjsmith@gmail.com}
            \and
            Johan Källström, Fredrik Heintz  \at
              Linköping University\\
              Linköping, Sweden\\
              \email{johan.kallstrom@liu.se}
              \email{fredrik.heintz@liu.se}
            \and
             Gabriel Ramos \at
              Universidade do Vale do Rio dos Sinos\\
              S\~ao Leopoldo, RS, Brazil\\
              \email{gdoramos@unisinos.br}
            \and
            Roxana R\u{a}dulescu \at
              AI Lab, Vrije Universiteit Brussel\\
              Belgium\\
              \email{roxana.radulescu@vub.be}
            \and
            Diederik M.\ Roijers \at
              Vrije Universiteit Brussel, Belgium  \& \\
              HU University of Applied Sciences Utrecht,
              the Netherlands\\
              \email{diederik.roijers@vub.be}
              \and
            Conor F. Hayes, Patrick Mannion \at
              National University of Ireland Galway\\
              Galway, Ireland\\
              \email{c.hayes13@nuigalway.ie}
              \email{patrickmannion@nuigalway.ie}
            \and
              Pieter J.K. Libin \at
              Vrije Universiteit Brussel, Belgium  \&\\
              Universiteit Hasselt, Belgium \& \\
              Katholieke Universiteit Leuven, Belgium\\
              \email{pieter.libin@vub.be}
            \and
              Richard Dazeley \at
              Deakin University, Geelong, Australia\\
              \email{richard.dazeley@deakin.edu.au}
}
\date{Received: date / Accepted: date}

\maketitle

\begin{abstract}
The recent paper ``Reward is Enough" by Silver, Singh, Precup and Sutton posits that the concept of reward maximisation is sufficient to underpin all intelligence, both natural and artificial. We contest the underlying assumption of Silver et al. that such reward can be scalar-valued. In this paper we explain why scalar rewards are insufficient to account for some aspects of both biological and computational intelligence, and argue in favour of explicitly multi-objective models of reward maximisation. Furthermore, we contend that even if scalar reward functions can trigger intelligent behaviour in specific cases, it is still undesirable to use this approach for the development of artificial general intelligence due to unacceptable risks of unsafe or unethical behaviour.

\keywords{Scalar rewards \and Vector rewards \and Artificial general intelligence \and Reinforcement learning \and Multi-objective decision making \and Multi-objective reinforcement learning \and Safe and Ethical AI}
\end{abstract}

\section{Introduction}
\label{sec:introduction}

Recently, Silver et al. \cite{silver2021reward} posited that the concept of reward maximisation is sufficient to underpin all intelligence. Specifically they present the \emph{reward-is-enough} hypothesis that ``Intelligence, and its associated abilities, can be understood as subserving the maximisation of reward by an agent acting in its environment", and argue in favour of reward maximisation as a pathway to the creation of artificial general intelligence (AGI). While others have criticised this hypothesis and the subsequent claims \cite{kilcher2021explained,ouellet2021notefficient,roitblat2021notenough,shead2021deepmind}, here we make the argument that Silver et al. have erred in focusing on the maximisation of scalar rewards. The ability to consider multiple conflicting objectives is a critical aspect of both natural and artificial intelligence, and one which will not necessarily arise or be adequately addressed by  maximising a scalar reward. In addition, even if the maximisation of a scalar reward is sufficient to support the emergence of AGI, we contend that this approach is undesirable as it greatly increases the likelihood of adverse outcomes resulting from the deployment of that AGI. Therefore, we advocate that a more appropriate model of intelligence should explicitly consider multiple objectives via the use of vector-valued rewards.

Our paper starts by confirming that the reward-is-enough hypothesis is indeed referring specifically to scalar rather than vector rewards (Section 2). In Section 3 we then consider limitations of scalar rewards compared to vector rewards, and review the list of intelligent abilities proposed by Silver et al. to determine which of these exhibit multi-objective characteristics. Section 4 identifies multi-objective aspects of natural intelligence (animal and human). Section 5 considers the possibility of vector rewards being internally derived by an agent in response to a global scalar reward. Section 6 reviews the relationship between scalar rewards, artificial general intelligence (AGI), and AI safety and ethics, before providing our proposal for a multi-objective approach to the development and deployment of AGI. Finally Section 7 summarises our arguments and provides concluding thoughts.

\section{Does the reward-is-enough hypothesis refer to scalar rewards?}

Before we argue against the use of scalar rewards, we first establish that this is in fact what Silver et al. are advocating. While the wording of the \emph{reward-is-enough} hypothesis as quoted above does not explicitly state that the reward is scalar, this is specified in Section 2.4 (``A reward is a special scalar observation $R_t$'') where the authors also state that a scalar reward is suitable for representing a variety of goals or considerations which an intelligent agent may display:

\begin{quote}
   A wide variety of goals can be represented by rewards. For example, \emph{a scalar reward signal can represent weighted combinations of objectives}, different trade-offs over time, and risk-seeking or risk-averse utilities. \cite[p.4, emphasis added]{silver2021reward}
\end{quote}

In addition, the authors later acknowledge the existence of other forms of reinforcement learning such as multi-objective RL or risk-sensitive methods, but dismiss these as being solutions to specialised cases, and contend that more general solutions (i.e. those based on maximising a cumulative scalar reward) are to be preferred. We will present an argument contesting this view in Section \ref{sec:scalar-limitations}.

\begin{quote}
    Rather than maximising a generic objective defined by cumulative reward, the goal is often formulated separately for different cases: for example multi-objective learning, risk-sensitive objectives, or objectives that are specified by a human-in-the-loop. [\textellipsis] While this may be appropriate for specific applications, a solution to a specialised problem does not usually generalise; in contrast a solution to the general problem will also provide a solution for any special cases. \cite[p.11]{silver2021reward}
\end{quote}

\section{The limitations of scalar rewards}
\label{sec:scalar-limitations}

\subsection{Theoretical limitations of scalar rewards}
The limitations of scalar rewards and the advantages of vector-based multi-objective rewards for computational agents have been extensively established in prior work \cite{hayes2021practical,ruadulescu2020multi,roijers2013survey}. In the interests of space and brevity, we focus here on the aspects that are of most relevance to the reward-is-enough hypothesis.

Clearly, many of the tasks faced by an intelligent decision-maker require trade-offs to be made between multiple conflicting objectives. For example a biological agent must aim to satisfy multiple drives such as reproduction, hunger, thirst, avoidance of pain, following social norms, and so on. A computational agent does not have the same physical or emotional motivations and so, when applied in the context of a highly-constrained task such as playing a board game like Go, there may be a single, clearly defined objective. However, it is likely that the agent will need to account for multiple factors in its decision making, when applied in less restricted environments. The ubiquity of multiple objectives is evident even in the cases presented by Silver et al. For example, in Section 3 they suggest that a squirrel's reward may be to maximise survival time, or reproductive success, or to minimise pain, while a kitchen robot may maximise healthy eating by its user, or their positive feedback, or some measure of the user's endorphin levels. An agent based on scalar rewards must either be maximising only one of these competing objectives, or some scalarised combination of them.

The prevalence of genuinely multi-objective tasks in the real-world is reflected in the thriving nature of research fields such as multi-criteria decision-making \cite{clemen1996making,triantaphyllou2000multi,velasquez2013analysis} and multi-objective optimisation \cite{coello2007evolutionary,deb2014multi}. Furthermore, this is reflected in the broad range of application areas that involve multi-objective aspects, as identified in \cite{coello2004applications,hayes2021practical}, which span almost all aspects of human society. In particular, any decision which affects multiple stake-holders will require consideration of multiple objectives \cite{coyle2020explaining}.

Silver et al. acknowledge that multiple objectives can exist, but state that ``a scalar reward signal can represent weighted combinations of objectives". In the context of multiple objectives, the agent is concerned with maximising some measure of \emph{utility} which captures the desired trade-offs between those objectives. While it is certainly true that rewards representing multiple objectives can be combined into a scalar value via a linear weighting, it is well known that this places limitations on the solutions which can be found \cite{das1997closer,vamplew2008limitations}\footnote{Specifically, there may be solutions which lie in concave regions of the Pareto front representing optimal trade-offs between objectives, and no linear weighting exists which will favour these particular solutions. For a practical example of the implications of this, see \cite{brys2013behaviour}.}. This means that a scalar representation of reward may not be adequate to enable an agent to maximise its true utility \cite{roijers2013survey}. In particular, some forms of utility such as lexicographic ordering cannot be represented as a scalar value \cite{debreu1997preferences}. In contrast, intelligence based on vector rewards and approaches that are explicitly multi-objective can directly optimise any desired measure of utility \cite{hayes2021practical}. 

A second advantage of multi-objective reward representations is that they allow for a greater degree of flexibility in adapting to changes in goals or utility. An agent using a multi-objective representation can follow behaviour which is optimal with respect to its current goal, while simultaneously performing learning with regards to other possible future goals or utility preferences. This allows for rapid or even immediate adaptation should the agent's goals or utility change, which we would argue is likely to arise in dynamic environments, particularly in the context of life-long learning. This approach, which is known in multi-objective reinforcement learning research as \emph{multi-policy learning} \cite{roijers2013survey}, cannot readily be achieved by an agent observing only a scalar reward. 

Finally we wish to address Silver et al.'s comment that ``a solution to a specialised problem does not usually generalise; in contrast a solution to the general problem will also provide a solution for any special cases". We disagree with the implied assumption that maximisation of a cumulative scalar reward is the general case. Scalar rewards (where the number of rewards $n=1$) are a subset of vector rewards (where the number of rewards $n\geq1$). Therefore, intelligence developed to operate in the context of multiple rewards is also applicable to situations with a single scalar reward, whereas the inverse is not true. A similar argument can be made regarding the generality of risk-aware decision-making compared to single-objective decision-making.

\subsection{The multi-objective nature of intelligent abilities}

Section 3 of Silver et al. identifies the following set of intelligent abilities which they assert could arise by maximising a scalar reward:  Knowledge and learning, Perception, Social intelligence, Language, Generalisation, Imitation, and General intelligence.

For some abilities such as knowledge and learning, and imitation, we concur with the reward-is-enough hypothesis. While multi-objective rewards may provide benefits relating to these areas such as improving efficiency, they are not strictly required in order for an intelligent agent to develop these abilities. However, we contend that other abilities are clearly multi-objective in nature. In the following subsections we will address the benefits which multi-objective rewards may provide over scalar rewards with regards to each of these aspects of intelligence. We believe that the issue of general intelligence merits a deeper discussion, particularly in regard to the creation of artificial general intelligence, and so we will defer discussion of this facet of intelligence until Section \ref{sec:general-intelligence}.

\subsubsection{Perception}

When discussing perception, Silver et al. note that there may be costs associated with carrying out actions to acquire information. As such, there is an implicit trade-off between the costs associated with misperception and the costs incurred in information gathering. Furthermore, there is no reason to assume that the relationship between these costs must be linear or that the relationship will remain fixed over time.

\subsubsection{Social intelligence}

Silver et al. identify that social intelligence may arise from reward maximisation in an environment populated by multiple agents, by simply letting an agent treat other agents as part of the environment. However, learning in multi-agent systems, and the emergence of social intelligence, is more naturally expressed as a multi-objective decision-making problem. In a competitive setting, the agent should consider its main goals as well as the long-term impact an action has on the future choices of its opponents, which may require a trade-off between the two categories of objectives. In a cooperative setting, agents may have different capabilities, needs, and rewards. Therefore, a solution must be found that represents a trade-off among objectives acceptable to all agents, such that it allows for the cooperation to be established and maintained over time. Having a clear idea of the utility will help deciding whether to maintain or break partnerships when utility changes. 

In multi-agent settings, the difference between vector-valued rewards -- in multi-agent settings often called payoffs -- and scalar-valued rewards is especially clear. For single-objective problems solutions can often be guaranteed, while in the multi-objective problem setting such guarantees cannot be obtained. This is even the case when the utility functions of the agents participating in the system are known upfront and are common knowledge. For example, while for single-criterion coalition formation games individually stable and even core stable partitions are guaranteed to exist, Igarashi et al.~\cite{igarashi2017multi} show that even for \emph{multi-criteria coalition formation games (MC2FGs)} with known \emph{linear} utility functions this is not necessarily the case. Furthermore, while for single objective normal form games Nash equilibria are guaranteed to exist, this is not necessarily the case for \emph{multi-objective normal form games (MONFGs)}~\cite{ruadulescu2020utility}. These considerations demonstrate that the multi-objective multi-agent problem setting is fundamentally different from its single-objective counterpart.

\subsubsection{Language}
By supporting communication of information and coordination of actions, language is clearly a beneficial attribute for an intelligent agent operating within a multi-agent environment. Prior work has demonstrated that agents that are able to communicate can achieve mutually beneficial cooperative behaviour, which may not be possible without communication \cite{das2019tarmac}. It has also been shown that reward-maximising agents can develop their own linguistic structures that exhibit advanced features such as compositionality and variability \cite{havrylov2017emergence}. In this regard the arguments of Silver et al. are therefore correct; maximisation of a scalar reward is sufficient to give rise to the development and use of language.

However we contend that scalar rewards do not suffice to account for the full complexity of language displayed by humans. The use of language is intertwined with the role of social intelligence, and so the arguments from the previous section also apply to the development of linguistic capabilities. Harari \cite{harari2014sapiens} describes how the development of a social language was a principle driver of the cognitive revolution separating modern humans from earlier humans and animals. 
While many animals can communicate factual information, threat warnings, and even lie, there is no evidence of communicating for the purpose of higher levels of intentionality associated with human social and cultural behaviours \cite{griffin1976animalawareness,dorothy1990monkeys,dennett1983intentional}. 

Language plays multiple roles in human interactions; in addition to communication of factual information, it can also serve to strengthen relationships, display emotions or personality, persuade or mislead others, etc. As such the use of language by an intelligent agent may be driven by a variety of different conflicting factors -- for example, an agent may wish to persuade another agent to carry out a particular action, while still preserving the long-term trust relationship with them. We argue that such sophisticated use of language can only emerge where there is a desire to go beyond simply achieving a particular reward and it is, therefore, far more likely to derive from a multi-objective approach to decision-making.

\subsubsection{Generalisation}
Silver et al. define generalisation as the ability required of an agent that is maximising its cumulative reward within ongoing interaction with a single complex environment. The agent will inevitably encounter situations which deviate from its past experience, and so to behave effectively in those novel situations it must be able to appropriately generalise from prior experience.

An agent maximising a scalar reward may exhibit some aspects of generalisation, such as generalising across variations in state. However other aspects of generalisation, such as generalising to new tasks, will be problematic. As the agent has only observed its current reward signal, it has no basis for adapting its behaviour should the reward signal undergo a significant change. In contrast, as discussed earlier, a multi-policy multi-objective agent can learn with regards to all components of the vector reward they are receiving, regardless of the extent to which each component contributes to their current utility. Should the agent's utility function change, it can almost immediately adapt its behaviour so as to optimise this new utility  \cite{vamplew2017steering,abels2019dynamic}.

\section{Multi-objective reinforcement learning in natural intelligences}

If our arguments in favour of multi-objective representations of reward are correct, then it would be expected that naturally evolved intelligences such as those in humans and animals would exhibit evidence of vector-valued rewards. In fact, evolution has developed organisms that delegate learning not just into multiple objectives but even into multiple learning systems within an organism. There are multiple objectives at a basic biological regulatory level, and these are matched with multiple objectives at every level of analysis of the organism. In this section we show that these cannot be reduced to any single objective.

\subsection{Distinguishing biological evolution and biological intelligence}

\begin{figure}
    \centering
    \includegraphics[width=1\columnwidth]{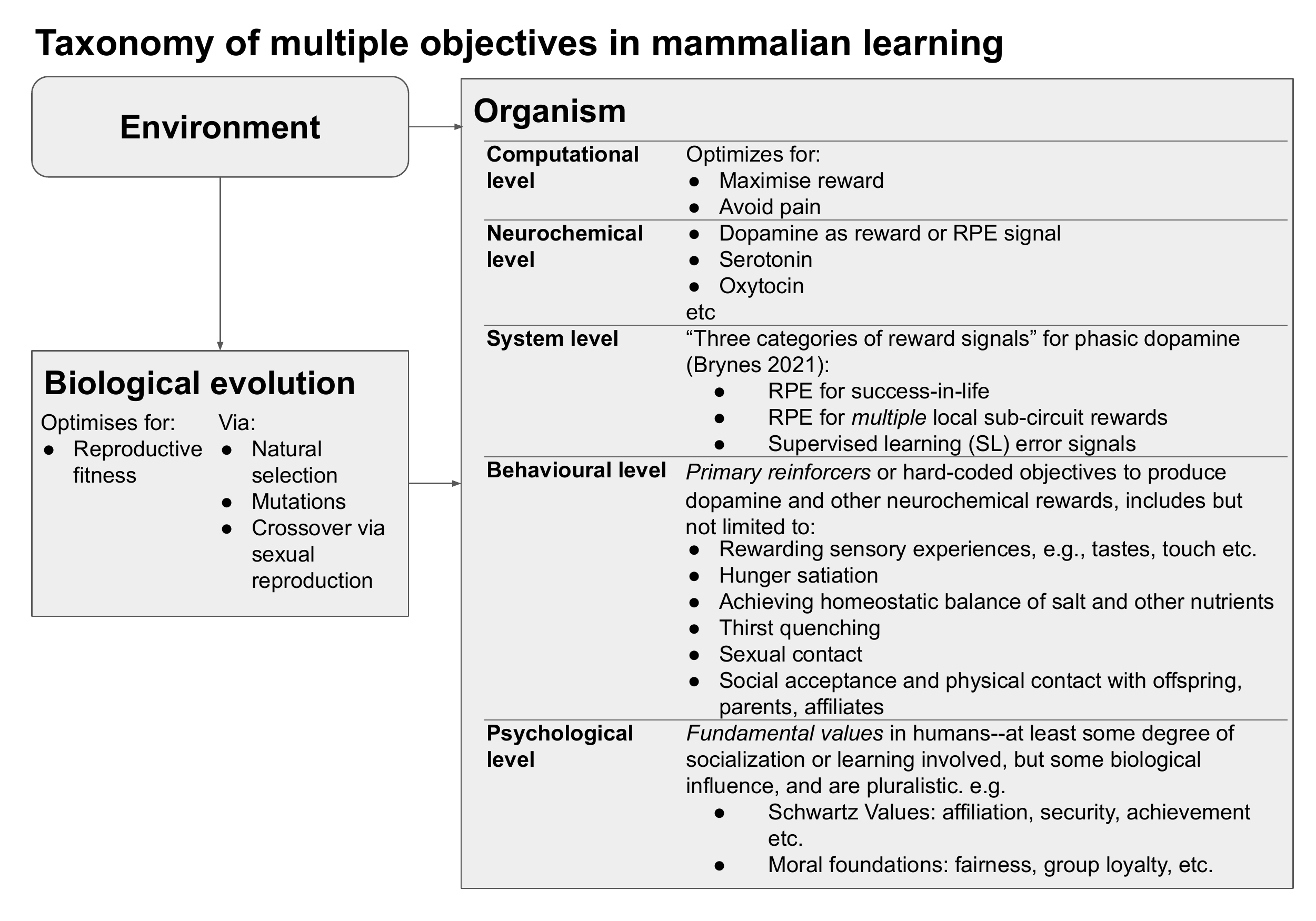}
    \caption{Biological evolution generates organism genotypes, optimising for the fit of an organism's phenotype to its environment. Genotypes are the genetic instruction set that are decoded into into an organism that exhibits a set of phenotypical expressions. These include all of the learning and reward systems that make up an organism's biological intelligence. 
    Further details about each level  of objectives in the organism are described in the text.
    }
    \label{fig:envs}
\end{figure}

When considering natural intelligences it is important to draw a clear distinction between the evolutionary process which has produced these intelligences, and the functioning of the intelligences themselves. Biological evolution optimises for a single objective (i.e. reproductive fitness), to an environment that varies over time and space (though its operationalisation depends greatly on the environment \cite{henrich2015secret,silver2021reward}). In contrast, at at least five distinct levels of analysis for human organisms, an organism's computational processes include multiple objectives that must be balanced during action selection (see Figure~\ref{fig:envs}).

\subsection{Biological intelligence has no single primary objective}

In optimising for reproductive fitness, evolution generates an organism's genotype (Figure~\ref{fig:envs}), which in turn creates what we could call an \emph{intellectual phenotype}, a set of innate intellectual capacities that an organism can use to learn about and interact with its environment. In mammals, as in most other organisms, the bio-computational processes that constitute that intellectual phenotype have no single objective; rather, they include multiple objectives including hunger satiation \cite{delgado2009reward}, thirst quenching, social bonding \cite{perret2015social}, and sexual contact \cite{fleischman2016evolutionary}, as described in Figure~\ref{fig:envs}.

Even if reproduction is regarded as a `single objective' of the evolutionary process, broadly construed, at the organism level it is not a primary reinforcer at all. Rather, the organism's phenotype includes a set of features, including intelligent systems, which have been tuned by the evolutionary process because they tend to lead to environmental fitness and ultimately genetic reproduction. This includes sexual orgasm \cite{fleischman2016evolutionary}, social contact with conspecifics like offspring \cite{perret2015social}, and so on. These do not necessarily internally store any kind of explicit representation of reproduction as a goal. Rather reproduction is an emergent result of the organism fulfilling other multiple primary objectives. Understanding the multi-objective nature of motivation in biological systems is critical because different systems are more dominant at different times, depending on context. An example in this regard concerns the remediation of pain and hunger. On the one hand, pain warrants an immediate response, while on the other hand, for hunger this response can be delayed for hours up to days. These examples demonstrate that such objectives could not be aggregated in a single reward signal, due to the difference in time scale in which these rewards are relevant, and we note that distinct biological subsystems are in place to support these differences. Simply understanding the computational processes of the organism is insufficient to predict behaviour without also including an account of the state of the organism and its environment, and the relevant objectives that arise as a function of that state.

The clearest example of biological primary objectives might be what are called `primary reinforcers' in behavioural psychology. These constitute behavioural goals that are innate \cite{delgado2009reward}, such as hunger satiation, thirst quenching, and sexual satisfaction (Figure~\ref{fig:envs}, `behavioural level'). These are themselves entire families of objectives, because an organism needs a wide array of nutrients to survive and have developed very specific objectives to ensure that each nutrient is obtained. Even for specific primary reinforcers, there may be multiple biological signals acting as proxy objectives to ensure those objectives are obtained. For instance, salt consumption alone activates taste receptors, interoceptive signals related to ingestion, and blood osmolality detectors designed to maintain homeostatic balance \cite{smith2021multiattributemodel}. These receptors drive the value assigned to consuming salt during decision-making at any particular moment. Thus, even for just a single critical nutrient, there are multiple biological objectives \cite{oka2013high,wolf1984multiple} on multiple level tuned to ensure an appropriate level of consumption.


At a psychological level (Figure~\ref{fig:envs}), humans appear to hold multiple irreducible and irreconcilable moral \cite{haidt2001emotional} and life \cite{SCHWARTZ2004230} objectives. Moral objectives include preferences for equality, for harm avoidance, and upholding authority hierarchies and group loyalties \cite{haidt2001emotional} and may have a basis in distinct biological tendencies. Schwartz et al. \cite{SCHWARTZ2004230} identified ten distinct human values, including benevolence, security, hedonism, and autonomy that appear to be distinct life objectives for people across cultures to varying degrees.

\subsection{There's no plausible neurochemical or neuroanatomical single objective function}

Proponents of a single-objective account of RL in biological organisms might look for a single neurotransmitter release that could conceivably underlie all the objectives outlined above -- perhaps a global reward signal or brain region that combines all the objectives outlined in the previous section into a single process. However, even if such a mechanism was identified, the reward-is-enough hypothesis would not follow: that single signal is properly thought of as \emph{hard-coded} to achieve multiple objectives, as outlined in Figure~\ref{fig:envs}.

Many of the biological and psychological objectives outlined above produce dopamine release when objectives are unexpectedly achieved; this is often thought of as a fundamental reward prediction error (RPE) signal \cite{mollick2020systems} (Figure~\ref{fig:envs}, `Neurochemical level'). This role is distinct from a global reward signal; dopamine seems to function as brain's RPE, rather than operationalising reward \cite{schultz1997neural}. Thus, an expected reward does not produce dopamine; dopamine is only released on the event of unexpected reward. It might be said only a single neurotransmitter performs the RPE function in mammalian reward learning, but this isn't the same as saying mammalian RL is single-objective, because that RPE signal is delivered to drive learning across a wide variety of domains. Other neurochemicals like oxytocin \cite{love2014oxytocin} or serotonin \cite{weng2013modulation} are important in the experience of pleasure, attachment, and motivation, but none could plausibly represent a universal reward signal involved in all goal-directed behaviour \cite{schultz1997neural,love2014oxytocin,weng2013modulation}.

A number of neuroanatomical brain regions are important for the production and assessment of value calculation and reward (Figure~\ref{fig:envs}, `Computational level'). A full survey of biological decision-making is outside of the scope of this article. But in brief, subcortical regions like the nucleus accumbens are thought to be involved in reward processing \cite{ikemoto1999role}, while the ventromedial prefrontal cortex (vmPFC) appears to formulate value signals associated with potential rewards. 
However, these do not appear to function as parts of single-objective reinforcement learning systems because the values represented are context dependent \cite{rudorf2014interactions}. When an organism is hungry, the value of food, as recorded in the vmPFC, is higher \cite{thomas2015satiation}; when an organism is thirsty, the value of drinking is higher. Something like a `common currency' might indeed be represented in the vmPFC \cite{levy2012root} and related regions, but to extend the metaphor, the `exchange' rate between that currency and various physiological and psychological goals and drives changes currently based on context, limiting the applicability of any single-objective account.

\subsection{The brain has multiple objective functions at a systemic level}

Byrnes \cite{byrnesphasicdopamine} attempts a description of the brain as an integrated reward learning system. The model described draws on Mollick et al.'s \cite{mollick2020systems} description of the brain's phasic dopamine system as well as classic work describing the brain as an array of parallel loops \cite{alexander1986parallel}. Here, three separate categories of phasic dopamine signals are described: reward prediction error for basic universal goals, reward prediction error for motor action execution, and supervised learning error signals. These are distinct processes required for human intelligence, all with their own objectives, but they are all required for a human brain (or a mammalian brain more generally) to function correctly.

\section{Internally-derived rewards}

One could argue that an agent concerned with maximising a scalar reward may still develop the capabilities required to carry out multi-objective decision-making. Natural intelligence provides an example of this. As we discussed in Section 4, the evolutionary objective of reproduction has led to the development of organisms with specialised sensors, internally derived reward signals, and learning systems associated with those signals. Another example of this is the perception of fairness and inequality, which has been identified as a process embedded in the human brain \cite{Cappelen15368}. Conceivably this could also arise in the context of computational intelligence, where agents based on evolutionary algorithms or reinforcement learning might construct their own internal reward signals to guide their learning and decision-making \cite{elfwing2008co,singh2010intrinsically,uchibe2008finding}.

One potential benefit of internalised rewards is that this may provide a less sparse reward signal. If the agent learns to identify environmental states or events which are correlated with future occurrences of its external reward, then creating secondary reward signals associated with those events may provide more immediate feedback, thereby speeding up learning and adaptation. For example, developing taste sensors which respond to particular nutrients in food will provide immediate rewards to an animal which eats that food, correlating to the more delayed benefits which may accrue from the intake of those nutrients. From the perspective of computational intelligence, it has been shown that multi-objectivisation (where the main scalar reward is decomposed into several distinct rewards that are correlated with the primary reward) can be beneficial for a computational RL agent \cite{brys2017multi,kurniawan2021}. Similarly, agents may use internally derived rewards to drive aspects of the learning process itself such as exploration \cite{barto2013intrinsic,oudeyer2009intrinsic}.

Regardless of whether an agent's reward vector is derived externally or internally, or through a  combination of both, the agent still requires the capacity to make decisions based on those vector values. Silver et al. argue that a computational agent maximising a single scalar reward could theoretically develop multi-objective capabilities. However this would require the agent to modify its own internal structure. Therefore, we believe that it is more practical to construct multi-objective agents through the explicit design of multi-objective algorithms. Similarly, we argue that where we can design multi-objective reward structures for computational agents, it makes sense to do so rather than to require them to identify such structures themselves. In fact, we contend that it typically will be easier to specify multi-objective rewards directly than to design a scalar reward which captures all of the various factors of interest. 

\section{Reward maximisation and general intelligence}\label{sec:general-intelligence}

\subsection{The risks of single-objective general intelligence}

One of the main arguments presented by Silver et al. is that the maximisation of even a simple reward in the context of a suitably complex environment (such as those which occur in the natural world) may suffice for the emergence of general intelligence. They illustrate this via the following scenario: 

\begin{quote}
  For example, consider a signal that provides +1 reward to the agent each time a round-shaped pebble is collected. In order to maximise this reward signal effectively, an agent may need to classify pebbles, to manipulate pebbles, to navigate to pebble beaches, to store pebbles, to understand waves and tides and their effect on pebble distribution, to persuade people to help collect pebbles, to use tools and vehicles to collect greater quantities, to quarry and shape new pebbles, to discover and build new technologies for collecting pebbles, or to build a corporation that collects pebbles. \cite[p.10]{silver2021reward}
\end{quote}

While the development of open-ended, far-reaching intelligence from such a simple reward is presented positively by Silver et al., this scenario is strikingly similar to the infamous \emph{paperclip maximiser}  thought experiment from the AI safety literature \cite{bostrom2003ethical}. In this example, a superintelligent agent charged with maximising the production of paperclips enslaves humanity and consumes all available resources in order to achieve this aim. While unrestricted maximisation of a single reward may indeed result in the development of complex, intelligent behaviour, it is also inherently dangerous \cite{omohundro2008}. For this reason, AI safety researchers have argued in favour of approaches based on satisficing rather than unbounded maximisation \cite{taylor2016}, or on multi-objective measures of utility which account for factors such as safety or ethics \cite{vamplew2018}.

Even when safety is not at risk, it is vital to carefully consider which types of behaviour can arise when deploying learning agents in society. Depending on the application, elements such as bluffing or information hiding can potentially be harmful and undesirable. For example, consider a smart grid setting, in which autonomous agents decide on behalf of the homeowners how to best store or trade locally generated green energy. This is, in general, a competitive scenario. While the main goal of each agent is to ensure the comfort and reduce the costs of the household, it is not acceptable for agents to manipulate the market or submit bluff-bids, even if these behaviours would be optimal with respect to their reward signal. Since autonomous agents need to operate in the context of our society, we should give careful thought to and investigate what type of emergent behaviour is desirable. This will be important to avoid the development of selfish and harmful mechanisms, under the pretext of being optimal with respect to a single numerical feedback signal.

While safety and ethics are not the focus of Silver et al.'s paper, it is concerning that these issues are not acknowledged in a paper which is actively calling for the development of AGI. More broadly we note that Silver et al. do not provide guidance as to the actual objective or reward signal which may be suitable for the creation of human-aligned AGI.

Even if we were to accept the hypothesis that an intelligent agent's behaviour can arise from maximising a single scalar reward within the environment, that does not necessarily imply that the design and construction of such a scalar reward is feasible.
For example, interpreting the behaviour of a squirrel as maximising a single scalar reward representing "survival" does not help to understand the mechanisms of that reward, nor does it assist in the construction of an equivalent reward signal that induces squirrel-like behaviour in an arbitrary intelligent agent.

As described by Amodei et al~\cite{amodei2016concrete}, reward specification is difficult 
even in trivial systems, and reward misspecification and reward hacking often lead to surprising, unintended, and undesirable behaviour. In more complex systems with more general agents, the potential for reward misspecification is significantly increased \cite{dewey2014reinforcement}. We argue, then, that the application of a single scalar reward signal leads to significant risk of unpredictable behaviour, and that even if a behaviour or intelligence can theoretically emerge from such a reward signal, predictable reward design is better achieved using multi-objective methods.

\subsection{A multi-objective approach to general intelligence}

In order for general artificial intelligence to be beneficial to humanity, it will need to be accountable and adaptable to human ethics, as well as human needs and aims. To be a part of society, an agent needs to adapt, and be accountable to others in this society, and we argue that agents that optimise for a single objective are severely handicapped in doing so. In this section, we will first explain the need for multiple objectives from an ethical and human-alignment point of view, and second explain our vision for future AI systems that form an integral part of society; agents that can be reviewed with respect to, and adapted to, a changing society.

Ultimately, all ethical capabilities that AGI can attain will have to come from humans, as well as all other goals and aims. There simply is no other available source for ethics and general goals than humans. However we do not agree amongst ourselves what ethically optimal behaviour is. Philosophers -- arguably the closest there is to generalist experts in ethics -- disagree on a wide range of ethical questions including meta-ethics, moral motivation, and normative ethics \cite{bourget2014philosophers}, and their positions on these disagreements correlate systematically with personal identities including race and gender. Psychologically, moral intuitions seem to arise from a pluralistic set of incommensurable and innate moral values that differ systematically across different political parties \cite{graham2013moral}. Fundamental human psychological values differ individually and across cultures \cite{SCHWARTZ2004230}. Furthermore, even if it is not a question of ethics, the question of what to care about is not trivial either \cite{frankfurt1982importance}. Therefore, we cannot expect people who need to specify the rewards for an AI system to get it right, especially not in one go. Moreover, priorities are likely to change over time. Any AI system -- general or not -- that is deployed for a long time, and that is possibly propagated to new generations of the systems, must be able to deal with this. 

Humans who will have to specify what the agent must care about will likely identify multiple goals that are important. For example, a self-driving car will need to minimise travel time, minimise energy consumption, as well as minimise the expected harm resulting from a trip to both the occupants of the car, other humans in the environment, animals, and property. Furthermore, a reasonable trade-off between these objectives (which may well be a non-linear one, as we explained in Section 3.1) must be identified during training. However, the engineers of such a system are most likely not ethicists nor legal experts. Furthermore, they do not own the utility of the system's deployment, and may not ultimately bear the responsibility for the system being deployed in practice. Therefore, the human-alignment process will necessarily be in the hands of others, who  need a clear explanation from the AI system about the available trade-offs between objectives in different situations. We therefore believe that an explicitly multi-objective approach is necessary for responsible agent design. For further arguments why human-aligned AI is inherently multi-objective, please refer to \cite{vamplew2018}.

Once the design, training, and testing is completed, and the AI system is deployed, it will be equipped with a set of objectives, and a mechanism (e.g., a utility function \cite{roijers2017multi}) for making online decisions about trade-offs between these objectives. However, it may well encounter situations that were not foreseen in training, and the trade-offs between objectives in these situations become unlike hitherto encountered trade-offs. Here we encounter another key benefit of an explicitly multi-objective approach -- and in our opinion an ethically necessary capability, of any agent. In combination with probabilistic methods, the agent can identify when it becomes too uncertain about which trade-offs will be desired and therefore, if possible, defer the decision back to responsible humans, or shut down safely. When this is not possible, these situations can be identified and reported back. For example, if a parrot suddenly gets loose in an automated factory, and the AI system correctly identifies it as an animal, but has never had to make choices between animal safety, human safety, product damage, and production before, the system may try to opt for safe system shutdown if a responsible human cannot be reached. However, if the parrot flies straight into the production line, more drastic immediate action might be required, and there is no time to shut down safely. These immediate actions will necessarily be taken on the fly, and will have to be reported and later reviewed to see whether the taken actions were indeed what the responsible humans would have wanted the system to do. 

\begin{figure}
    \centering
    \includegraphics[width=\textwidth]{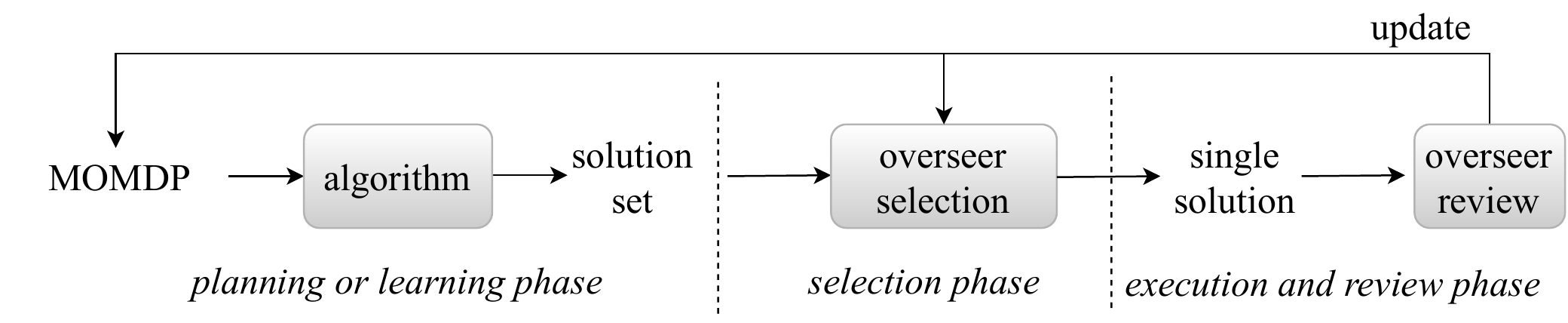}
    \caption{Our proposal for a responsible review and adjust scheme for future AI. When deployed, a new situation or undesirable outcome with respect to human preferences may lead to an adjustment in overseer selection (without a need to remodel or retrain), while errors in value estimates or new situations that lead to high uncertainty in the value estimates of one or more objectives may lead to additional training without remodelling, while the identification of a new objective may lead to changing the system itself.
    }
    \label{fig:review}
\end{figure}

Finally, when undesirable things have occurred, an agent needs to be able to explain the decision made. Single objective systems are only able to provide simple details such as what was the perceived state and that its chosen action maximised its reward \cite{cruz2019memory}. 
Such explanations provide little understanding to the users. However, an explicitly multi-objective approach confers significant further benefits. 
Namely, it can help diagnose exactly what went wrong, such as: what was the trade-off between objectives; what alternative outcomes would have occurred with alternative trade-offs; or, was the selected policy providing an undesired trade-off between objectives \cite{dazeley2021levels}.
Additionally, a multi-objective approach allows us to utilise a review-and-adjust cycle shown in Figure \ref{fig:review}, allowing an agent, possibly with direct or indirect user feedback, to automatically select a Pareto-optimal policy. Once selected, the reason for that selection can also be explained in the event of undesirable outcome. 
Hence, a multi-objective approach allows explicitly attributed details, as well as contrastive and counterfactual explanations of the reasoning process behind behaviour rather than just the outcome \cite{dazeley2021explainable}. Being able to explicitly identify the trade-off underlying an agent's basis for reasoning has long been argued as a key component of transparency in AGI \cite{hibbard2008open}. 

The review and adjust scheme \cite{hayes2021practical} has three major phases. During the planning or learning phase, an AGI would utilise a multi-objective reinforcement learning or planning algorithm to compute a set of optimal policies for all possible utility functions. In the selection phase, a policy that best reflects the AGI's utility function is chosen. The selected policy is then executed, during the execution phase. The outcome from the policy execution can then be reviewed by the overseer (either a human, the AGI itself, or another AGI). The MOMDP, utility function or set of solutions can then be updated based on the outcome from the previously executed policy and the review of those outcomes. We note that such reviews can not only be triggered by incidents, but also by regular inspection.

We see such a review-and-adjust cycle as an essential feature of future AI. We, as AI researchers have to enable responsible deployment, and well-informed review of the systems we create is a key feature of this. It is our opinion that the above-mentioned benefits are not merely desirable, but that it is a moral imperative for AI designers to obtain them, in order to create AI systems that more likely benefit society. In addition to the mathematical, technical, and biological arguments for why scalar reward is not enough, we thus also point out that there are ethical and societal reasons why scalar rewards are not enough. 

We acknowledge the difficulties which may arise in implementing human oversight of AGI if the latter achieves superhuman levels of intelligence \cite{bostrom2014superintelligence,everitt2018agi,alfonseca2021superintelligence}. Superintelligent AGI may be motivated to, and highly capable of, deceiving human overseers, or its behaviours and reasoning may simply be too complex for human understanding. Nevertheless we would argue that it is certainly preferable to attempt such oversight than not, and that a multi-objective AGI will provide greater transparency than a single-objective AGI. Nevertheless, we acknowledge that such an approach will not necessarily guarantee a manageable superhuman AGI,  and therefore a careful ethical consideration of such research efforts is warranted.

\section{Conclusion}

Silver et al. argue that maximisation of a scalar reward signal is a sufficient basis to explain all observed properties of natural intelligence, and to support the construction of artificial general intelligence. However, this approach requires representing all of the different objectives of an intelligence as a single scalar value. As we have shown, this places restrictions on the behaviour which can emerge from maximisation of this reward. Therefore, we contend that the \emph{reward-is-enough} hypothesis does not provide a sufficient basis for understanding all aspects of naturally occurring intelligence, nor for the creation of computational agents with broad capabilities.

In the context of the creation of AGI, a focus on maximising scalar rewards creates an unacceptable exposure to risks of unsafe or unethical behaviour by the AGI agents. This is particularly concerning given that Silver et al. are highly influential researchers and employed at DeepMind, one of the organisations best equipped to expand the frontiers of AGI. While Silver et al. ``hope that other researchers will join us on our quest", we instead hope that the creation of AGI based on reward maximisation is tempered by other researchers with an understanding of the issues of AI safety \cite{krakovna2020avoiding,leike2017ai} and an appreciation of the benefits of multi-objective agents \cite{abdolmaleki2020distributional,abdolmaleki2021multi}.

\begin{acknowledgements}
This research was supported by funding from the Flemish Government under the ``Onderzoeksprogramma Artifici\"{e}le Intelligentie (AI) Vlaanderen'' program, 
and by the National Cancer Institute of the U.S. National Institutes of Health under Award Number 1R01CA240452-01A1. 
The content is solely the responsibility of the authors and does not necessarily represent the official views of the National Institutes of Health or of other funders.

Pieter J.K. Libin acknowledges support from the Research Foundation Flanders (FWO, fwo.be) (postdoctoral fellowship 1242021N). 

Johan K\"allstr\"om and Fredrik Heintz were partially supported by the Swedish Governmental Agency for Innovation Systems (grant NFFP7/2017-04885), and the Wallenberg Artificial Intelligence, Autonomous Systems and Software Program (WASP) funded by the Knut and Alice Wallenberg Foundation.

Conor F. Hayes is funded by the National University of Ireland Galway Hardiman Scholarship.

\end{acknowledgements}



\vskip 0.2in
\bibliography{main.bib}
\bibliographystyle{spmpsci}

\end{document}